\RequirePackage[table]{xcolor}
\documentclass[journal,twoside,web]{ieeecolor}
\usepackage{tmi}
\usepackage{iftex}
\ifXeTeX\usepackage[T1]{fontenc}\fi
\usepackage{cite}
\usepackage{amsmath,amssymb,amsfonts}
\usepackage{graphicx}
\usepackage{booktabs}
\usepackage{multirow}
\usepackage{hyperref}
\hypersetup{hidelinks}
\usepackage{textcomp}
\usepackage{etoolbox}
\usepackage{stfloats}
\usepackage{cuted}
\makeatletter
\newenvironment{analysisfigure*}{%
  \begin{strip}\begin{minipage}{\textwidth}\centering\sf
  \def\@captype{figure}%
}{\end{minipage}\end{strip}}
\newenvironment{analysisfigure}{%
  \par\addvspace{\intextsep}\noindent
  \begin{minipage}{\columnwidth}\centering\sf\def\@captype{figure}%
}{\end{minipage}\par\addvspace{\intextsep}}
\makeatother
\usepackage[above,below]{placeins}

\providecommand{\refname}{References}
\patchcmd{\thebibliography}{\section*{\color{black}}}{\section*{\refname}}{}{\errmessage{refname patch failed}}

\definecolor{oursrow}{HTML}{F5EFF4}

\newcommand{\venue}[1]{\,{\tiny\textcolor{black!65}{#1}}}

\makeatletter
\let\ps@titlepagestyle\ps@plain
\makeatother
\begin{document}
\bstctlcite{BSTcontrol}
\title{SlimFormer++: Token Mixing Where It Matters for Efficient 3D Medical Image Segmentation}
\author{Yang Hong, Lei Zhang, Zhihong Liu, Jianqing Mo, Xujiong Ye
\thanks{(Corresponding author: Lei Zhang.)}
\thanks{Yang Hong, Lei Zhang, Zhihong Liu, and Xujiong Ye are with the University of Exeter, Exeter, U.K. (e-mail: yh693@exeter.ac.uk; L.Zhang6@exeter.ac.uk; zl645@exeter.ac.uk; X.Ye2@exeter.ac.uk).}
\thanks{Jianqing Mo is with Guangdong University of Technology (e-mail: momolon@gdut.edu.cn).}}

\maketitle

\begin{abstract}
To determine where computation is needed in volumetric segmentation networks, we analyze block-level redundancy and stage-wise token-mixer contribution. Centered Kernel Alignment reveals high input--output representational similarity within blocks of representative convolutional, hybrid, and Transformer-style segmentation networks. Similarity-guided pruning reduces their parameter counts by 39--75\% while decreasing mean Dice by less than 0.4 percentage points. To examine token mixers within blocks, we introduce Anatomical Relation Distortion, which quantifies how bypassing a mixer distorts feature relations between labeled anatomical positions. Applied to an unpruned candidate architecture, Centered Kernel Alignment indicates high representational redundancy in intermediate encoder mixers, whereas Anatomical Relation Distortion shows that they contribute little and the bottleneck mixer most. These findings motivate SlimFormer++, which uses Selective Mixing blocks with windowed Mamba mixing in the first encoder stage and only residual channel MLPs in deeper stages. At the bottleneck, Augmented Mixing injects similarity-weighted sparse long-range context into the axial taps of a depthwise convolution at cost linear in voxel count. Evaluated against 12 baselines across eight public CT, MRI, and ultrasound datasets (1,809 cases), SlimFormer++ achieves the highest mean Dice (87.84\%) and lowest mean HD95 (6.50~mm) with the fewest parameters (5.98 million). These results support stage-aware token mixing as a design strategy for improving the accuracy--efficiency trade-off in volumetric segmentation.
The source code is available at \url{https://github.com/WHHHHHY822/SlimFormerPlusPlus}.
\end{abstract}

\begin{IEEEkeywords}
Efficient network, medical image segmentation, redundancy analysis, token mixer, Transformer
\end{IEEEkeywords}

\begin{figure}[!tb]
    \centering
    \includegraphics[width=\columnwidth]{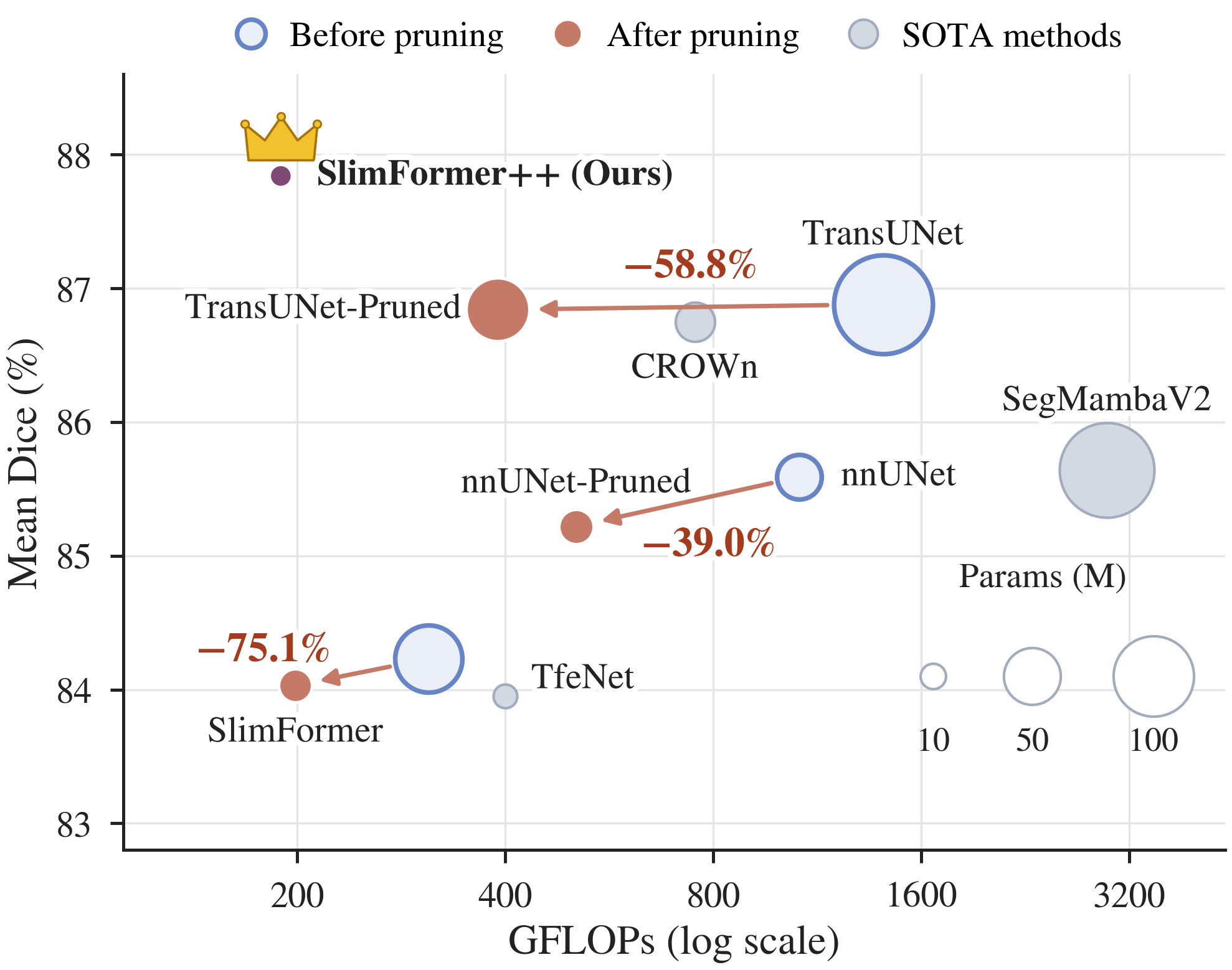}
    \caption{Performance--efficiency trade-off of representative 3D segmentation architectures.
    Dice is the unweighted mean over the eight datasets. Arrows connect each model before pruning (hollow) to its pruned variant (filled), and the percentage on each arrow is the resulting parameter reduction.}
    \label{fig.1}
\end{figure}

\section{Introduction}\label{sec 1}

\IEEEPARstart{E}{fficient} volumetric medical image segmentation has received increasing attention because clinical deployment often requires accurate three-dimensional delineation under strict constraints on memory, latency, and computational cost \cite{jeon2025no}. High spatial resolution and large through-plane depth make dense feature extraction and long-range context modeling costly in CT and MRI volumes. Recent segmentation architectures \cite{isensee2021nnu,chen2024transunet,xing2025segmamba} therefore typically adopt U-shaped encoder--decoder designs in which repeated computational units progressively encode, transform, and fuse multiscale volumetric features. These units use convolution \cite{roy2023mednext,han2022convunext}, self-attention \cite{ntanzi2025unetr++,yu2023metaformer}, or state-space operations \cite{liu2024swin,liao2024lightm} to transform spatial features. Under the MetaFormer abstraction \cite{yu2023metaformer}, these spatial operations serve as token mixers within computational blocks that are stacked across stages to form a hierarchical encoder. The repeated application of these blocks motivates a fundamental architectural question: are all \textbf{blocks} necessary to maintain segmentation accuracy?

Lightweight operators \cite{chen2025frequency,jin2025slim,yang2025gated} reduce the computational burden of individual blocks, but they do not reveal whether the repeated use of such blocks across the network is necessary. Hybrid designs show the benefit of combining different operators \cite{song2026freqconvmamba,gan2026wtcm}, but leave open how the contribution of each operator varies across depth.

Using Centered Kernel Alignment (CKA) \cite{davari2023reliability}, we examine block-level redundancy in convolutional nnUNet, hybrid TransUNet, and the Transformer-style SlimFormer backbone \cite{hong2025slimformer}. Similarity-guided pruning reduces model size with little loss of segmentation accuracy (Fig.~\ref{fig.1}). Because block-level pruning keeps or removes each block as a whole, every retained block keeps its token mixer. This leaves a finer-grained question in architecture design: Does the \textbf{token mixer within each block} contribute equally at every stage?

We therefore examine token mixing along two complementary axes: representation redundancy measured by CKA and contribution measured by Anatomical Relation Distortion (ARD), as illustrated in Fig.~\ref{fig.2}. ARD quantifies how bypassing a token mixer distorts feature relations between labeled anatomical positions. Across the tested mixers, token-mixing contribution is nonuniform across depth: it is highest at the bottleneck (BN), followed by the first encoder stage, and smaller in the remaining encoder stages. These findings support a stage-aware design principle: strong token mixing at the bottleneck, lightweight mixing at the first encoder stage, and none in the remaining encoder stages.

We instantiate this principle in SlimFormer++ (Fig.~\ref{fig.3}), using Selective Mixing (SM) in the encoder and Augmented Mixing (AM) at the bottleneck. Evaluation on eight public CT, MRI, and ultrasound datasets demonstrates its accuracy--efficiency trade-off.

Our main contributions are summarized as follows:
\begin{enumerate}
\item \textbf{Block-level redundancy analysis.} We use CKA to identify redundant blocks in convolutional, hybrid, and Transformer-style volumetric segmentation networks. Pruning these blocks reduces parameter counts by 39--75\% while lowering mean Dice by less than 0.4 percentage points (pp). These findings motivate the compact hybrid macro architecture of SlimFormer++, described in Section~\ref{sec 3.2}.

\item \textbf{Stage-aware architecture design.} We introduce ARD, a contribution measure complementary to CKA, and use the two measures to derive the stage-aware principle for token mixing. SlimFormer++ implements this principle through Selective Mixing, which restricts windowed Mamba mixing to the first encoder stage, and Augmented Mixing, which injects similarity-weighted sparse remote context into a bottleneck depthwise convolution at linear cost in voxel count (Section~\ref{sec 3.3}).

\item \textbf{Extensive evaluation.} We compare SlimFormer++ with state-of-the-art methods and pruned baselines on eight public datasets spanning CT, MRI, and ultrasound. SlimFormer++ achieves the highest mean Dice (87.84\%) and lowest mean HD95 (6.50 mm) among the compared methods. In the efficiency comparison, it also has the fewest parameters (5.98M), the lowest computational cost in GFLOPs, and the lowest peak inference memory.
\end{enumerate}

\section{Related Work}\label{sec 2}

\subsection{Efficient Medical Image Segmentation Architectures}

Encoder design is a central consideration in medical image segmentation architectures \cite{salpea2022medical,azad2024medical}. Traditional convolutional networks \cite{lee20233d,roy2023mednext,isensee2021nnu} aggregate local context, whereas Transformer-based models \cite{kamal2026adaptive,he2023swinunetr} and state-space models \cite{xing2025segmamba, gong2025nnmamba} capture long-range dependencies. Hybrid designs, including TransUNet \cite{chen2024transunet}, nnWNet3D \cite{zhou2025nnwnet}, CROWn \cite{huang2026crown}, and PnPNet \cite{you2026towards}, integrate convolutional processing with broader contextual interactions. Efficiency-oriented designs, such as 3D UNETR++ \cite{ntanzi2025unetr++}, LightM-UNet \cite{liao2024lightm}, and TfeNet \cite{WU2026103882}, reduce computational requirements through lightweight feature-extraction and mixing modules. However, reducing the cost of individual blocks does not by itself establish whether their spatial-mixing operations are necessary at each network stage.

Another line of research examines the contribution of the token mixer to model performance. MetaFormer \cite{yu2023metaformer} shows that the general Transformer architecture, rather than the specific token mixer, accounts for much of the performance: pooling, random, or even identity mixers remain competitive on image classification. SlimFormer \cite{hong2025slimformer} reaches a similar conclusion for 3D medical image segmentation, where replacing the token mixer changes accuracy only slightly. However, comparisons between separately trained networks do not directly quantify the contribution of token mixing at individual stages within a fixed trained model. Where operators do vary across depth, their placement follows design heuristics: the hybrid MetaFormer variants and CoAtNet \cite{dai2021coatnet} place pooling or convolution before attention, and SwinUNETR-V2 \cite{he2023swinunetr} adds stage-wise convolutions to a Swin Transformer encoder. SlimFormer++ measures mixer contribution within a trained network and uses its variation with depth to allocate token mixing.

\subsection{Representation Redundancy and Structured Pruning}

Layer-level redundancy has been widely reported in Transformer-based architectures, particularly in large language models (LLMs). ShortGPT \cite{men2025shortgpt} removes layers whose input and output hidden states are highly similar, Gromov \emph{et al.} \cite{gromov2025unreasonable} prune many deeper layers with little degradation, and He \emph{et al.} \cite{he2026uncovering} find that many attention layers can be dropped selectively. CKA \cite{davari2023reliability} measures representational similarity, which does not guarantee functional redundancy. Its scores can be manipulated without substantially altering model behavior \cite{davari2023reliability}, and similar representations can have different functional behavior \cite{klabunde2025similarity}. These limitations underscore the need to verify that removing a transformation preserves task performance. LLM layer-removal findings also need validation in multiscale segmentation networks with encoder--decoder hierarchies and skip connections.

Structured pruning improves computational efficiency by removing selected filters or channels, as in ThiNet \cite{luo2017thinet}. For medical image segmentation, SlimFormer \cite{hong2025slimformer} uses representational similarity to identify redundancy in encoder blocks and their token-mixing and MLP components. SlimFormer++ reformulates redundancy analysis as a problem of allocating token-mixing capacity across network depth. It uses CKA to analyze redundancy across three architectures and ARD to measure mixer contribution within each stage. These measures guide the simplification of encoder mixers and the concentration of mixing capacity at the bottleneck.

\section{Method}\label{sec 3}

\begin{figure*}[!tb]
    \centering
    \includegraphics[width=\textwidth,trim=9.34bp 1109.59bp 6.67bp 239.39bp,clip]{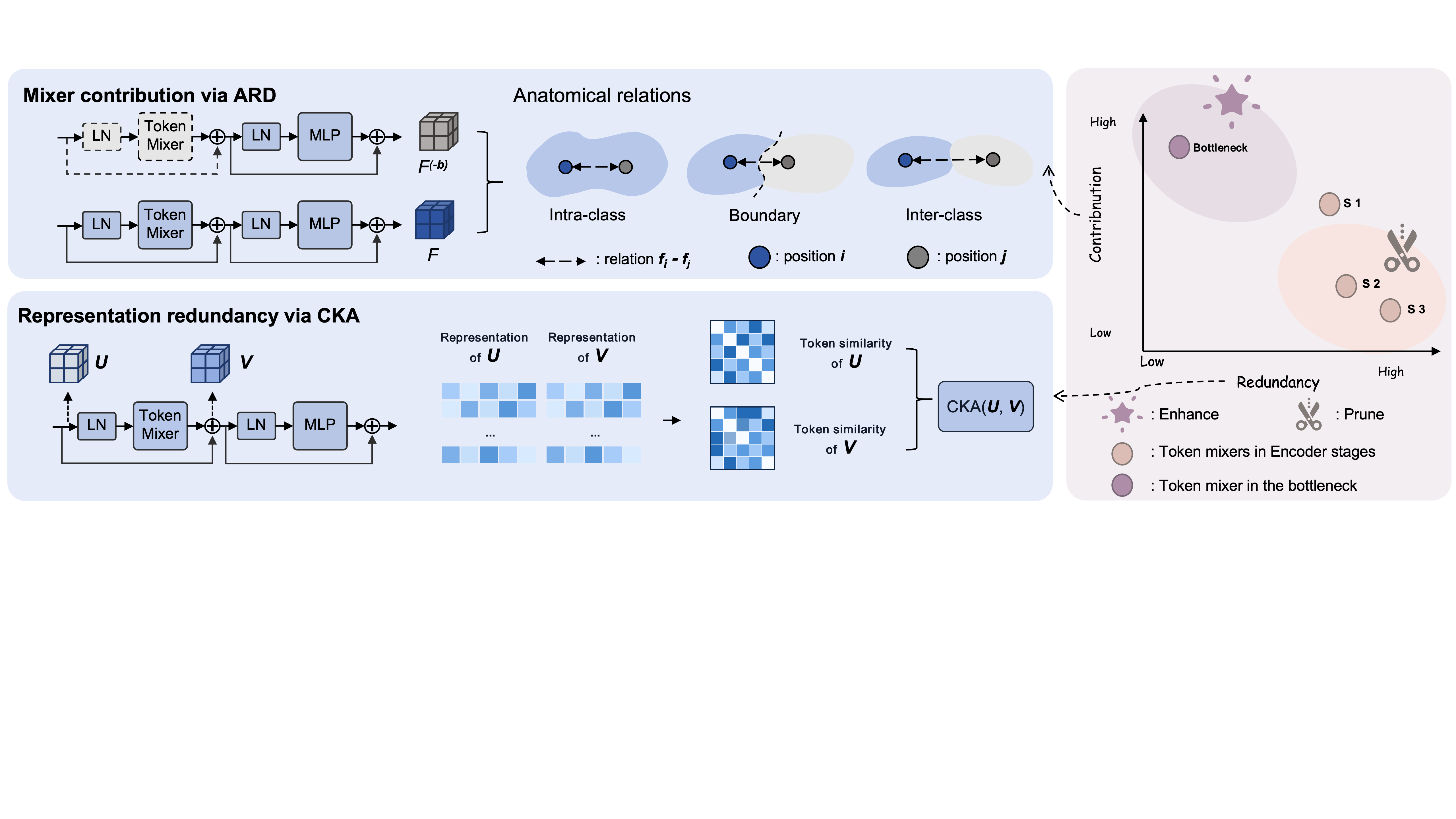}
    \caption{
    Redundancy and contribution measures for token mixers. ARD compares the intact features $\mathbf{F}$ with the features $\mathbf{F}^{(-b)}$ obtained by bypassing the token mixer of a block, and CKA compares the representations $\mathbf{U}$ and $\mathbf{V}$ before and after token mixing. The right panel sketches the stage-wise pattern measured on the unpruned network (Section~\ref{sec 4.3.2}).
    }
    \label{fig.2}
\end{figure*}

\subsection{Redundancy and Contribution Measures}
\label{sec 3.1}

We analyze redundancy in complete encoder blocks and the contribution of their token mixers. Block-level CKA in three representative architectures informs the macro design (Section~\ref{sec 3.2}). At the mixer level, CKA measures representational redundancy and ARD measures relation distortion under bypass, providing the two axes of Fig.~\ref{fig.2}. These measures guide the stage-wise allocation of token mixers (Section~\ref{sec 3.3}).

\textbf{Block formulation.} In Transformer-style architectures, block-level analysis treats each block as an input-output mapping. Mixer-level analysis separates token and channel mixing into two pre-normalized residual branches. For input $\mathbf{X}$ to block $b$,
\begin{equation}
\begin{aligned}
\mathbf{X}_{\mathrm{mix}} &= \mathbf{X}+g_b\,\mathcal{M}_b\left(\mathrm{LayerNorm}(\mathbf{X})\right),\\
\mathbf{X}_{\mathrm{out}} &= \mathbf{X}_{\mathrm{mix}}+\mathrm{MLP}\left(\mathrm{LayerNorm}(\mathbf{X}_{\mathrm{mix}})\right),
\end{aligned}
\label{eq:shared_mixer_block}
\end{equation}
where $\mathcal{M}_b$ mixes spatial positions, $\mathrm{MLP}$ mixes channels independently at each position, and $g_b\in\{0,1\}$ controls retention of the token-mixing branch. Setting $g_b=0$ removes that branch and leaves the residual channel MLP.

\textbf{Representation redundancy measured by CKA.}

For block- or stage-level analysis, let $\mathbf{U}_b\in\mathbb{R}^{n\times C_{\mathrm{in}}}$ and $\mathbf{V}_b\in\mathbb{R}^{n\times C_{\mathrm{out}}}$ denote the input and output representations of block $b$ (or of a whole stage) at $n$ sampled spatial positions, with the input max-pooled to the output grid when the block or stage downsamples; for a mixer block in \eqref{eq:shared_mixer_block}, they are $\mathbf{X}$ and $\mathbf{X}_{\mathrm{out}}$. For mixer-level analysis, they are $\mathbf{X}$ and $\mathbf{X}_{\mathrm{mix}}$, the representations before and after token mixing. Their linear CKA similarity is
\begin{equation}
S_{\mathrm{CKA}}(b)=
\frac{
\left\|\widetilde{\mathbf{U}}_b^{\top}\widetilde{\mathbf{V}}_b\right\|_F^2
}{
\left\|\widetilde{\mathbf{U}}_b^{\top}\widetilde{\mathbf{U}}_b\right\|_F
\left\|\widetilde{\mathbf{V}}_b^{\top}\widetilde{\mathbf{V}}_b\right\|_F
},
\end{equation}
where $\widetilde{\mathbf{U}}_b$ and $\widetilde{\mathbf{V}}_b$ are the centered representations. A high input--output CKA score indicates strong representational alignment and potential redundancy. Similarity-guided pruning therefore ranks the blocks of a trained network by $S_{\mathrm{CKA}}(b)$ and removes a subset of the highest-ranked ones without a fixed threshold. We apply this procedure to nnUNet, TransUNet, and SlimFormer~\cite{hong2025slimformer}; Section~\ref{sec 4.1.2} describes the training protocols, and Fig.~\ref{fig.1} shows the resulting parameter reductions. Because CKA aggregates alignment across organ and background regions, changes in discriminative feature directions or local anatomical relationships relevant to prediction may have little effect on the overall score.

\textbf{Token-mixer contribution measured by ARD.}
To measure token-mixer contribution to label-defined anatomical feature relations, we introduce Anatomical Relation Distortion (ARD), which compares two forward passes with fixed model parameters: the intact network and the same network with the mixer of the target block $b$ bypassed by setting $g_b=0$ in \eqref{eq:shared_mixer_block}. Let $\mathbf{F},\mathbf{F}^{(-b)}\in\mathbb{R}^{n\times C}$ denote the intact and bypassed feature matrices, respectively (Fig.~\ref{fig.2}). Their rows, $\mathbf{f}_i^\top$ and $\mathbf{f}_i^{(-b)\top}$, represent features at corresponding sampled spatial positions. These features are read after the residual channel MLP of block $b$. Using the same ground-truth-defined pair set $\mathcal{P}$ in both forward passes, we sample intra-class pairs with the same label, boundary pairs consisting of nearby positions across label boundaries, and inter-class pairs with different labels. These pairs assess within-class consistency, local boundary separability, and inter-class separation, respectively. For each pair $(i,j)\in\mathcal{P}$, stacking the within-pass differences $\mathbf{z}_{ij}=\mathbf{f}_i-\mathbf{f}_j$ and $\mathbf{z}_{ij}^{(-b)}=\mathbf{f}_i^{(-b)}-\mathbf{f}_j^{(-b)}$ yields the anatomical relation matrices $\mathbf{Z}$ and $\mathbf{Z}^{(-b)}$.

Let $\mu=\|\mathbf{Z}\|_F$, $\nu=\|\mathbf{Z}^{(-b)}\|_F$, and $\kappa=\|\mathbf{Z}^{\top}\mathbf{Z}^{(-b)}\|_*$, where $\|\cdot\|_F$ and $\|\cdot\|_*$ denote the Frobenius and nuclear norms. ARD measures how much these relations change under bypass after the channels of the two passes are aligned:
\begin{equation}
\begin{aligned}
D_{\mathrm{ARD}}(b)
&=\frac{\min_{\mathbf{O}^{\top}\mathbf{O}=\mathbf{I}}\left\|\mathbf{Z}-\mathbf{Z}^{(-b)}\mathbf{O}\right\|_F^2}{\mu^2+\nu^2}\\
&=1-\frac{2\kappa}{\mu^2+\nu^2},
\end{aligned}
\label{eq:ard}
\end{equation}
where $\mathbf{O}$ ranges over $C\times C$ orthogonal matrices, and the second line gives its closed form \cite{schonemann1966generalized}. For $\mu^2+\nu^2>0$, $D_{\mathrm{ARD}}$ lies in $[0,1]$ and is zero exactly when bypass changes the relations only by an orthogonal transformation of the channels. ARD captures changes in relation magnitude and geometry after orthogonal alignment. We use larger distortion under bypass as a proxy for greater mixer contribution and assess its relationship to segmentation accuracy in Section~\ref{sec 4.2}.

The stage-level score $D_{\mathrm{ARD}}(s)$ applies \eqref{eq:ard} with $g_b=0$ jointly for all blocks $b$ in stage $s$ (S1--S3 or BN), using the output features of stage $s$. To compare relative contribution profiles across networks and seeds, we normalize the scores within each trained network by the maximum score across its stages:
\begin{equation}
\rho_s=\frac{D_{\mathrm{ARD}}(s)}{\max_{s'} D_{\mathrm{ARD}}(s')}.
\label{eq:normalized_contribution}
\end{equation}

\begin{figure*}[!tb]
    \centering
    \includegraphics[width=\textwidth]{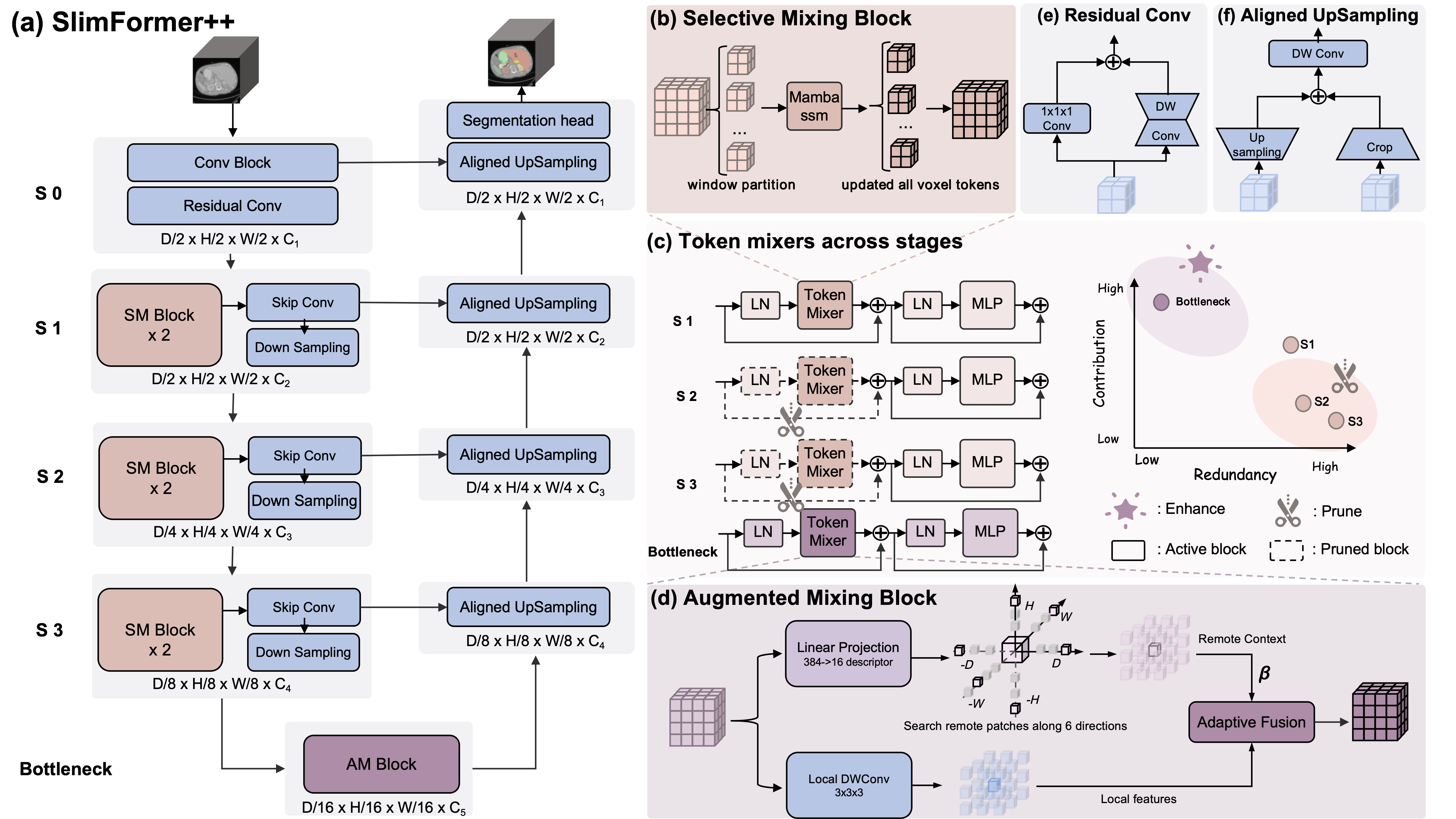}
    \caption{Overall framework of SlimFormer++. (a) Macro architecture: a convolutional Stage 0, three encoder stages (S1--S3), a bottleneck (BN), and a decoder. (b) Selective Mixing (SM) block: a causal Mamba scan within nonoverlapping windows. (c) Token mixers across stages, set by the redundancy--contribution analysis (right): active in S1, pruned in S2 and S3, where only the channel MLPs remain, and strengthened at the bottleneck (d). (d) Augmented Mixing (AM) block: remote context sampled along six axial directions is mixed into the axial taps of the depthwise convolution. (e) Residual convolution block of Stage 0. (f) Aligned upsampling module of the decoder.}
    \label{fig.3}
\end{figure*}

\subsection{Macro Architecture of SlimFormer++}
\label{sec 3.2}

Fig.~\ref{fig.3}(a) illustrates the overall architecture of SlimFormer++, which follows a hierarchical encoder--decoder architecture with skip connections. The SM and AM blocks are mixer blocks of the form in \eqref{eq:shared_mixer_block}, with $\mathcal{M}_b$ being the SM mixer $\mathcal{S}$ (Section~\ref{sec 3.3.1}) in the encoder and the AM mixer $\mathcal{A}$ (Section~\ref{sec 3.3.2}) at the bottleneck.

For an input volume of spatial size $(D\times H\times W)$, Stage 0 comprises a convolutional block followed by a residual convolutional block (Fig.~\ref{fig.3}(e)), yielding features of size $(\frac{D}{2}\times\frac{H}{2}\times\frac{W}{2}\times C_1)$. Motivated by the block-level redundancy analysis in Fig.~\ref{fig.1}, we use this shallow stem for local anatomical feature extraction and downsampling, reducing the token count for subsequent Transformer-style stages.

S1--S3 each contain two SM blocks (Fig.~\ref{fig.3}(b)), followed by a skip convolution, which provides representations for decoder fusion, and a downsampling layer. A single AM block (Fig.~\ref{fig.3}(d)) forms the bottleneck. Indexing S1--S3 and BN by $s=1,\ldots,4$, the SM/AM blocks in stage~$s$ operate on features downsampled by a factor of $2^s$ along each spatial dimension, with $C_{s+1}$ channels.

In the decoder, each aligned upsampling module (Fig.~\ref{fig.3}(f)) upsamples and crops the coarser features to match the corresponding skip representation, combines them through element-wise addition, and refines the fused features using depthwise convolution. A segmentation head produces the final prediction.

\subsection{Micro Architecture of Token Mixers}
\label{sec 3.3}

The micro design sets the gates $g_b$ in \eqref{eq:shared_mixer_block} stage by stage and specifies the Selective Mixing (SM) and Augmented Mixing (AM) mixers. Token mixing in S1 adds regional context to high-resolution local features, whereas bottleneck mixing relates distant locations in the input volume. S1 context propagates through the encoder, and decoder fusion combines the skip features of S2 and S3 with bottleneck context. These complementary paths motivate the expectation of limited additional mixer contributions in S2 and S3. Halving each spatial dimension reduces the token count eightfold between stages, so S1 favors lightweight mixing with cost linear in token count, whereas the bottleneck can afford stronger remote interaction.

\textbf{Mixer allocation.} For a trained unpruned network with SM in S1--S3, AM at the bottleneck, and $g_b=1$ in every block, we compute CKA and normalized stage-level ARD on validation cases (Section~\ref{sec 3.1}). The SlimFormer++ allocation is $g_b=1$ for blocks in S1 or BN and $g_b=0$ for blocks in S2 or S3 (Fig.~\ref{fig.3}(c)); the empirical profiles supporting this choice are reported in Section~\ref{sec 4.3.2}. The compact network inherits the retained weights and is fine-tuned.

\subsubsection{Selective Mixing in the Encoder}
\label{sec 3.3.1}

SM combines stage-wise branch selection through $g_b$ with windowed Mamba mixing. In stage~$s$, the mixer pads $\bar{\mathbf{X}}=\mathrm{LayerNorm}(\mathbf{X})$ to multiples of $P=8$ along each spatial dimension, partitions it into $P\times P\times P$ windows, and flattens each window in depth--height--width row-major order into $P^3$ tokens. A standard unidirectional Mamba module~\cite{gu2023mamba} processes each sequence, with its state reset to zero at the start of every window. The outputs are reassembled and cropped to give $\mathcal{S}(\cdot)$ in \eqref{eq:shared_mixer_block}.

With a fixed window size, SM has linear cost in voxel count and restricts token interactions to individual windows.

\subsubsection{Augmented Mixing at the Bottleneck}
\label{sec 3.3.2}

AM augments a depthwise convolution with sparse remote context weighted by feature similarity. At the bottleneck, adjacent voxels are 16 input voxels apart along each axis, so remote sampling spans large distances in the input volume. AM injects remote context into the axial taps of the depthwise kernel, allowing local and remote features to share kernel coefficients. AM thus expands contextual coverage beyond the local convolutional neighborhood, whereas sparse sampling and shared kernel coefficients limit the additional computational and parameter costs.

Let $\Omega$ denote the bottleneck grid with extents $L_d$, $L_h$, and $L_w$, and $N=|\Omega|$ voxels. Let $\mathbf{x}_i\in\mathbb{R}^{C}$ be the feature at voxel $i$ in $\mathrm{LayerNorm}(\mathbf{X})$, and $y^{\mathrm{dw}}_{i,c}$ the output of a $3\times3\times3$ depthwise convolution. AM augments its six axial taps $\mathcal{T}_{\mathrm{ax}}=\{\pm\mathbf{e}_d,\pm\mathbf{e}_h,\pm\mathbf{e}_w\}$, where $\mathbf{e}_a$ is the unit step along axis $a\in\{d,h,w\}$.

\textbf{Remote candidates.} For each tap $t=\sigma\mathbf{e}_a$, $\sigma\in\{-1,+1\}$, candidates lie on the same axis at radii proportional to its extent:
\begin{equation}
\begin{aligned}
r_{a,k}&=\max\!\left(2,\operatorname{round}\!\left(\tfrac{k}{4}(L_a-1)\right)\right),\\
\mathcal{N}_t(i)&=\left\{i+\sigma r_{a,k}\mathbf{e}_a\in\Omega\;:\;k\in\mathcal{K}_a\right\},
\end{aligned}
\end{equation}
where $\mathcal{K}_a\subseteq\{1,2,3,4\}$ retains the smallest slot index for each distinct radius not exceeding $L_a-1$; candidates outside $\Omega$ are masked.

\textbf{Similarity-weighted tap augmentation.} Descriptors $\boldsymbol{\phi}_i=\mathbf{W}_{\phi}\mathbf{x}_i\in\mathbb{R}^{d_{\phi}}$, $d_{\phi}=16$, score candidate relevance. For each nonempty $\mathcal{N}_t(i)$, we aggregate the original features:
\begin{equation}
\begin{aligned}
\alpha^t_{ij}&=\operatorname*{softmax}_{j\in\mathcal{N}_t(i)}\!\left(\frac{\cos(\boldsymbol{\phi}_i,\boldsymbol{\phi}_j)}{\tau}+\eta_{t,k(j)}\right),\\
\mathbf{m}_{i,t}&=\sum_{j\in\mathcal{N}_t(i)}\alpha^t_{ij}\,\mathbf{x}_j,
\end{aligned}
\end{equation}
where $\tau=0.2$ is the temperature, $k(j)$ indexes the candidate slot, and $\eta_{t,k}$ is a learnable direction- and slot-specific bias. The aggregated context augments the corresponding axial tap, yielding $\mathcal{A}(\cdot)$ in \eqref{eq:shared_mixer_block}:
\begin{equation}
y_{i,c}=y^{\mathrm{dw}}_{i,c}+\sum_{\substack{t\in\mathcal{T}_{\mathrm{ax}}\\\mathcal{N}_t(i)\neq\varnothing}}
K_{c,t}\,\beta_c\bigl(m_{i,t,c}-x_{i+t,c}\bigr),
\end{equation}
where $K_{c,t}$ is the depthwise kernel coefficient at tap $t$, and $\beta_c$ is a learnable coefficient shared within each of eight groups of 48 channels and initialized to 0.1.

\textbf{Properties and complexity.} AM reduces exactly to the depthwise convolution when all $\beta_c=0$ and adds approximately 6.2k parameters in our configuration. With $Q=2\sum_a|\mathcal{K}_a|\le24$ retained candidate slots, its cost is $\mathcal{O}(N[27C+Cd_{\phi}+Q(C+d_{\phi})])$. The fixed candidate budget makes this cost linear in voxel count, compared with $\mathcal{O}(N^2C)$ for dense self-attention.

\section{Experiments}
\label{sec 4}
We first examine whether CKA and ARD provide reliable guidance for architecture design (Section~\ref{sec 4.2}), then examine the macro and micro design (Section~\ref{sec 4.3}). Section~\ref{sec 4.4} compares mixer replacements. Sections~\ref{sec 4.5} and~\ref{sec 4.6} compare segmentation performance and computational cost with existing methods, and Section~\ref{sec 4.7} presents qualitative results.

\subsection{Experimental Setup}
\label{sec 4.1}

\subsubsection{Datasets}
\label{sec 4.1.1}

We evaluate SlimFormer++ on eight public 3D segmentation datasets comprising 1,809 cases and covering CT, MRI, and ultrasound across four body regions (Table~\ref{tab:dataset_summary}).

We use AbdomenCT-1K, ACDC, and AMOS for the stage-level analysis guiding token-mixer allocation (Section~\ref{sec 3.3}) and the mixer replacement experiments (Section~\ref{sec 4.4}).

\begin{table}[!htb]
\centering
\caption{Summary of the eight public segmentation datasets.}
\label{tab:dataset_summary}
\footnotesize
\setlength{\tabcolsep}{3pt}
\renewcommand{\arraystretch}{1.12}
\begin{tabular*}{\columnwidth}{@{\extracolsep{\fill}}lllc@{}}
\toprule
Dataset & Region & Target & Modality \\
\midrule
AbdomenCT-1K \cite{ma2022abdomenct1k}& Abdomen & Multi-organ & CT \\
ACDC \cite{bernard2018acdc}& Thorax & Ventricles & MRI \\
AMOS \cite{ji2022amos}& Abdomen & Multi-organ & CT \\
CHAOS \cite{kavur2021chaos}& Abdomen & Liver & MRI \\
RESECT \cite{xiao2017resect}& Head & Brain tumor & US \\
MMWHS \cite{zhuang2019mmwhs}& Thorax & Whole heart & CT, MRI \\
VESSEL12 \cite{rudyanto2014vessel12}& Thorax & Lungs & CT \\
Leg3DUS \cite{duque2024leg3dus}& Lower limb & Muscles & US \\
\midrule
Total & 4 regions & -- & 3 \\
\bottomrule
\end{tabular*}
\end{table}

\subsubsection{Implementation Details}
\label{sec 4.1.2}

We implement all methods in PyTorch within the nnU-Net v2 framework~\cite{isensee2021nnu}. For each dataset, all methods use the same publicly provided data splits, dataset-specific preprocessing, patch size, and nnU-Net v2 default data augmentation. Inputs are padded when required by an architecture, and predictions are cropped back to the original patch size.

The pruned nnUNet and TransUNet models are fine-tuned from the retained weights, whereas both the SlimFormer backbone and its adopted pruned architecture are retrained.

All models use stochastic gradient descent with Nesterov momentum $0.99$, an initial learning rate of $10^{-2}$, and weight decay $3\times10^{-5}$. The learning rate follows a polynomial decay schedule with exponent $0.9$. The loss equally weights soft Dice and cross-entropy. Deep supervision is disabled, and gradient norms are clipped to $12$. 

Experiments use PyTorch 2.6 and CUDA 12.6 on four NVIDIA GH200 GPUs, with distributed data parallelism for multi-GPU training and automatic mixed precision where supported. Checkpoints are selected by nnU-Net's exponential moving average of online foreground Dice. Inference uses sliding windows with $50\%$ overlap, Gaussian importance weighting, and mirroring-based test-time augmentation.

\subsubsection{Evaluation Metrics}
\label{sec 4.1.3}

We report the Dice similarity coefficient (Dice)~\cite{dice1945measures} and the 95th percentile Hausdorff distance (HD95)~\cite{kuijf2019standardized}. For the mixer replacement experiments, we additionally report the Jaccard index~\cite{jaccard1901etude} and average surface distance (ASD), computed symmetrically~\cite{heimann2009comparison}. Dice and Jaccard are reported in percent, and HD95 and ASD in millimeters. 

\subsection{Validity of CKA and ARD as Design Diagnostics}
\label{sec 4.2}

Fig.~\ref{fig.4} supports using CKA and ARD as complementary diagnostics to guide subsequent macro and micro design analyses. CKA assesses stage-level redundancy, whereas ARD assesses the contribution of each stage's token mixers. Higher CKA corresponds to smaller Dice decreases after stage removal, whereas higher ARD corresponds to larger decreases after mixer bypass.

\begin{samepage}
Stage removal skips an entire stage; mixer bypass disables only its token mixers, preserving residual paths and channel MLPs. Removing any encoder stage substantially reduces Dice, S1-S3 all show large Dice decreases following stage removal, despite differences in CKA. While BN has substantially higher CKA and a smaller Dice decrease. In contrast, mixer bypass produce the highest ARD and the largest Dice decrease at the bottleneck and the smallest at S2/S3. S1 shows negligible Dice loss, but its ARD exceeds those of S2/S3 by more than an order of magnitude. An encoder stage can therefore remain essential even when its token mixers contribute little.
\par
\end{samepage}

\begin{analysisfigure}
    \centering
    \includegraphics[width=\columnwidth]{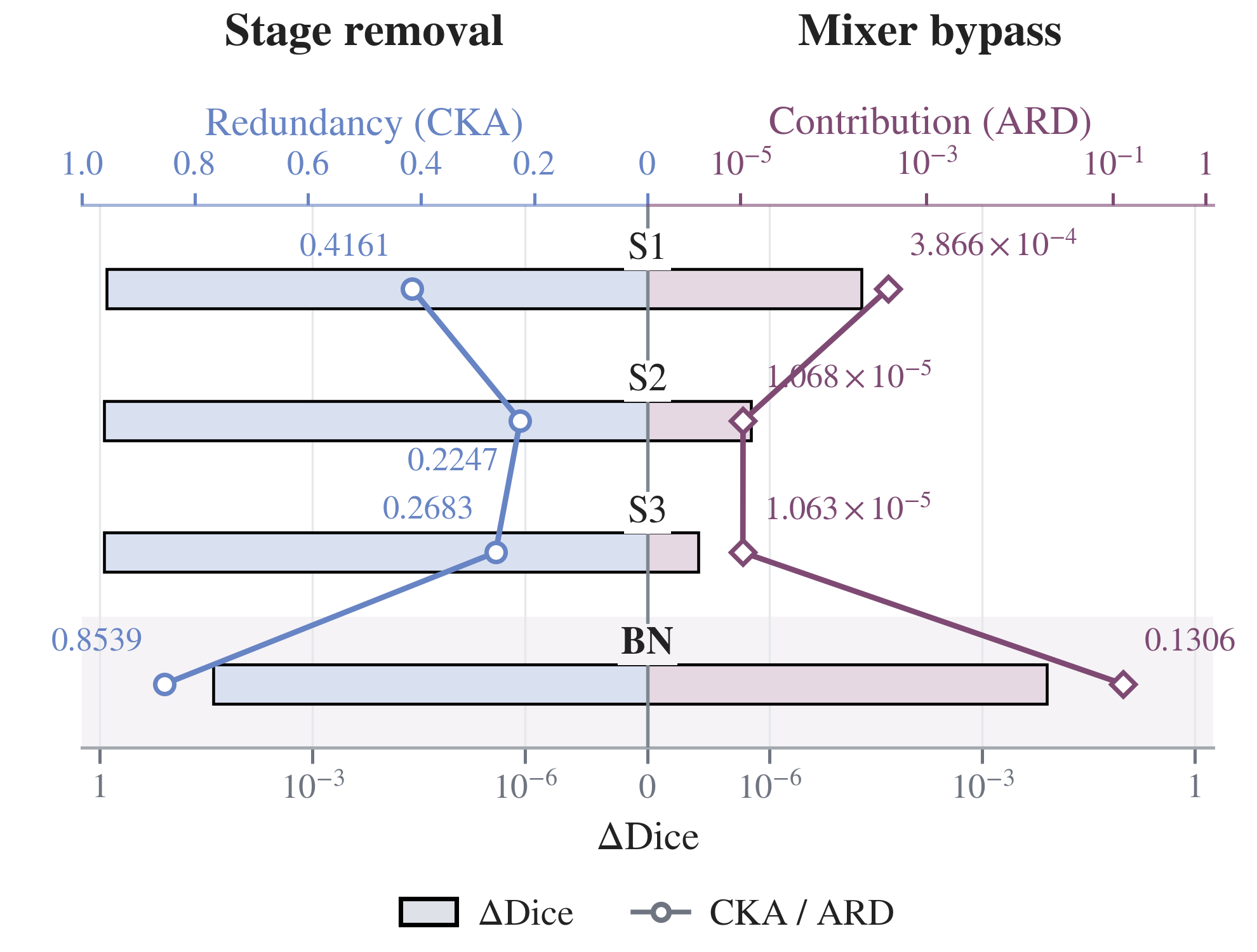}
    \caption{Sensitivity of the unpruned network to stage removal (left) and joint bypass of a stage's token mixers (right). Bars show the Dice drop ($\Delta$Dice, symmetric log scale); lines show whole-stage CKA and ARD. BN denotes the bottleneck.}
    \label{fig.4}
\end{analysisfigure}

\begin{analysisfigure*}
    \centering
    \includegraphics[width=0.9\textwidth]{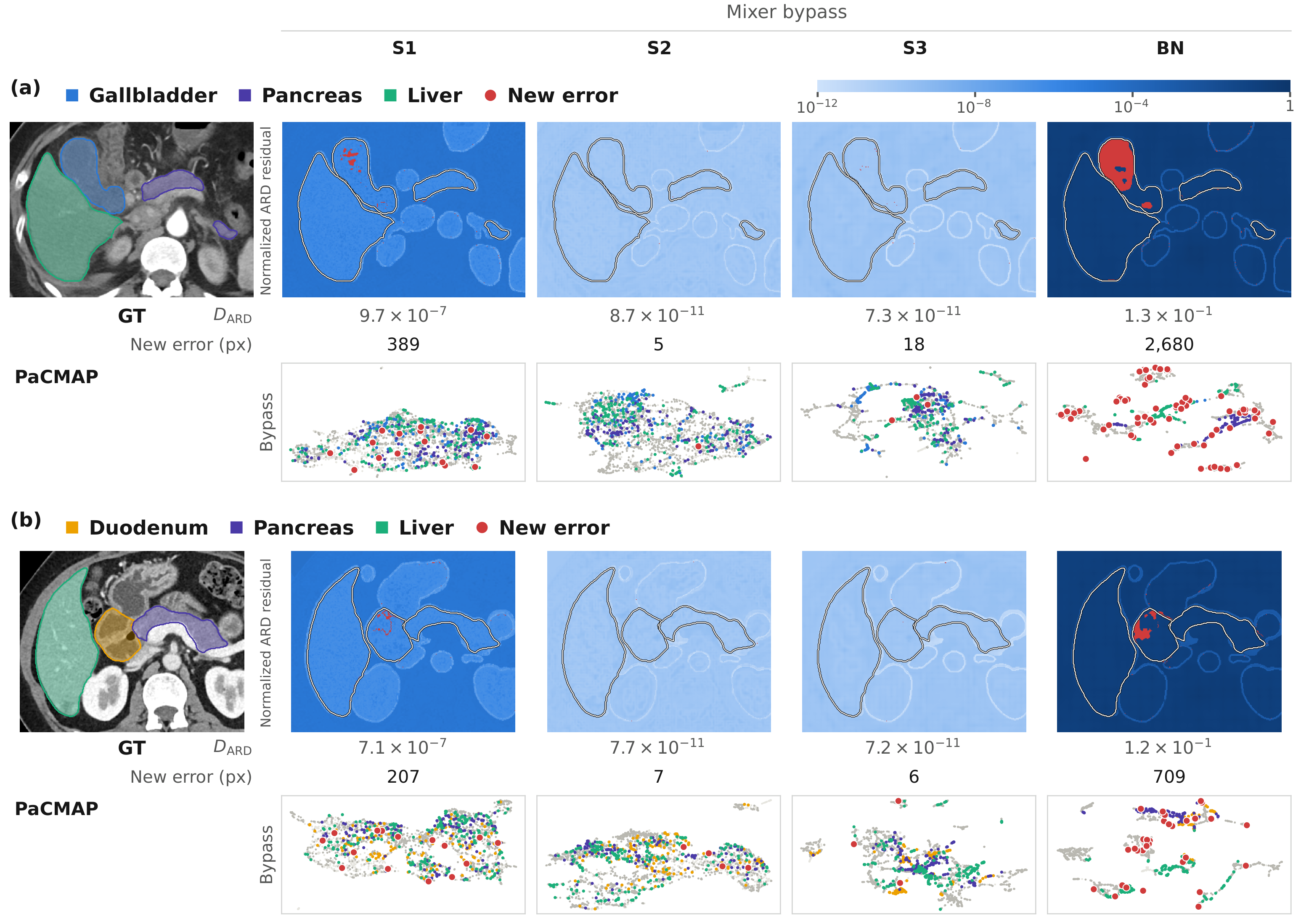}
    \caption{Token-mixer bypass on challenging classes: the gallbladder (a) and the duodenum (b), with pancreas and liver as references. Each column bypasses one stage's token mixers. Heatmaps show normalized ARD residuals, and PaCMAP shows the bypassed features. Red marks newly misclassified pixels and tokens, with pixel counts below each heatmap. BN denotes the bottleneck.}
    \label{fig.5}
\end{analysisfigure*}

\clearpage

\begin{samepage}
\textbf{Analysis of challenging classes.}
Aggregate Dice can mask losses on challenging classes through gains on easier ones. We therefore analyze the gallbladder and duodenum, with the pancreas and liver as references (Fig.~\ref{fig.5}). We examine the effects of stage-wise token-mixer bypass from two complementary perspectives: changes in segmentation results, shown by newly incorrect predictions alongside ARD residual heatmaps, and changes in feature space, visualized using PaCMAP~\cite{wang2021pacmap} embeddings of the bypassed features.
\par
\end{samepage}

Bypassing the bottleneck mixer produces widespread normalized ARD residuals ($D_{\mathrm{ARD}}\approx 0.1$), but new errors concentrate in challenging classes: 2,680 gallbladder pixels are misclassified as background and 709 duodenum pixels as stomach, whereas the liver remains largely correct. S1 bypass produces scattered errors ($D_{\mathrm{ARD}}\approx 10^{-6}$), whereas S2/S3 bypass yields at most 18 newly misclassified pixels in one slice ($D_{\mathrm{ARD}}\approx 10^{-10}$). PaCMAP~\cite{wang2021pacmap} shows the same ordering: misclassified tokens cluster densely after bypassing the bottleneck mixer, scatter after S1 bypass, and are nearly absent after S2/S3 bypass.

\subsection{Diagnostic Analysis of Macro and Micro Design}
\label{sec 4.3}

\subsubsection{Macro Design: Block-Level Redundancy}
\label{sec 4.3.1}

\begin{samepage}
We first examine block-level representational redundancy in the macro design of SlimFormer++. The compared networks differ in the number and composition of blocks per stage, so we compare them at the shared stage granularity. Fig.~\ref{fig.6} reports stage-averaged CKA, defined as the mean whole-stage input--output CKA across the encoder stages and bottleneck, excluding the stem.
\par
\end{samepage}

\begin{samepage}
Pruning lowers stage-averaged CKA in all three baselines, with the largest reduction in nnUNet and smaller changes in TransUNet and SlimFormer.
\par
\end{samepage}

Among the unpruned baselines, hybrid TransUNet has the lowest stage-averaged CKA. SlimFormer++ combines convolution with SlimFormer-style Transformer blocks in a newly designed architecture (Section~\ref{sec 3.2}). Its stage-averaged CKA of 0.441 is lower than those of unpruned TransUNet (0.463), nnUNet (0.517), and SlimFormer (0.594). This difference indicates lower whole-stage input--output similarity in SlimFormer++.

\begin{analysisfigure}
    \centering
    \includegraphics[width=\columnwidth]{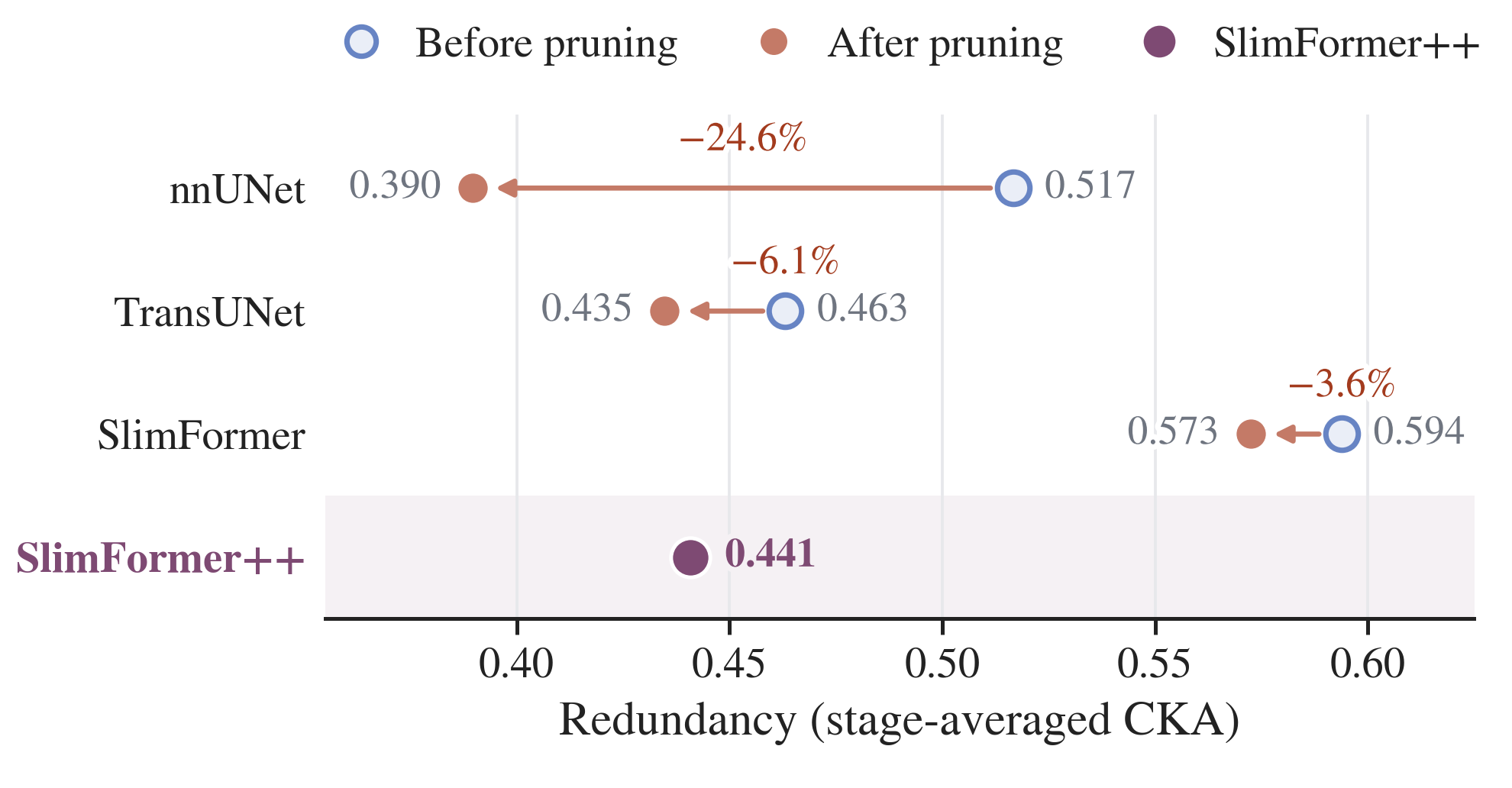}
    \caption{Representation redundancy is measured by stage-averaged CKA, the mean whole-stage input--output CKA over the encoder stages and the bottleneck. Higher CKA indicates greater representational similarity and potential redundancy. }
    \label{fig.6}
\end{analysisfigure}

\begin{analysisfigure*}
    \centering
    \includegraphics[width=\textwidth]{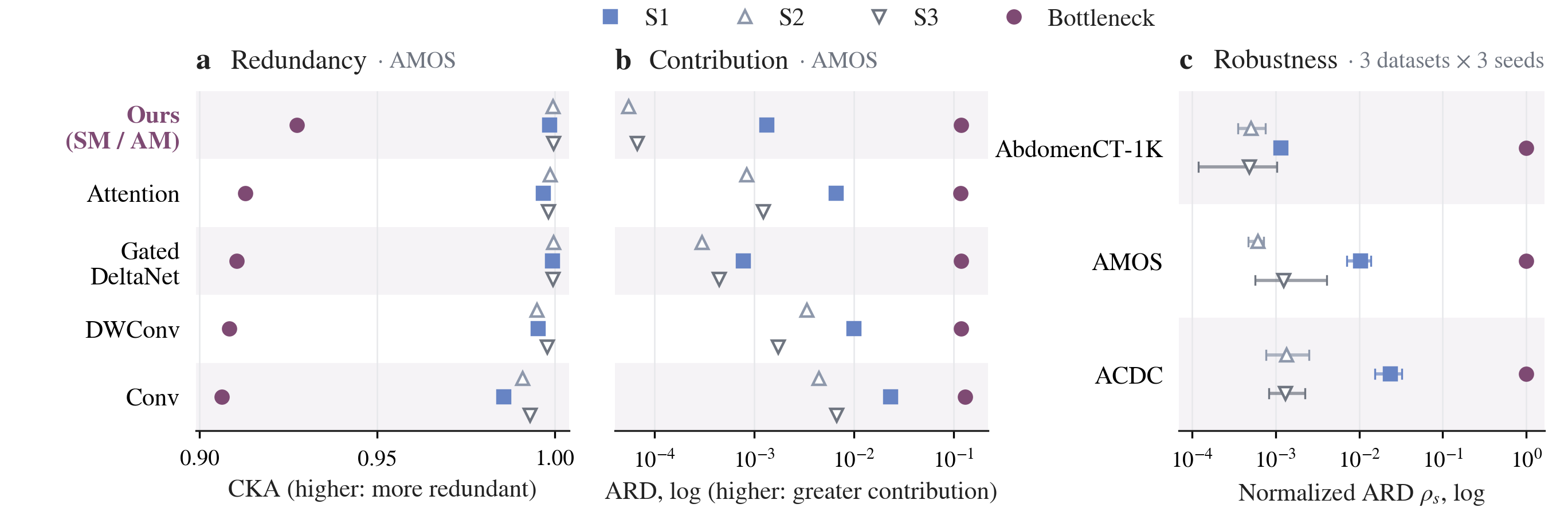}
    \caption{Stage-wise token-mixer diagnostics and robustness. All networks share the same skeleton except for token mixers. (a) Mixer-level CKA, averaged over the blocks within each stage, and (b) ARD under joint stage-wise mixer bypass on AMOS. (c) Normalized contribution $\rho_s$ in \eqref{eq:normalized_contribution} of the unpruned network across three datasets; points show geometric means and horizontal bars the min--max range over three training seeds.}
    \label{fig.7}
\end{analysisfigure*}

\subsubsection{Micro Design: Stage-Wise Mixer Contribution}
\label{sec 4.3.2}

Fig.~\ref{fig.7}(a,b) shows that the nonuniform, stage-wise contribution of token mixing is intrinsic to the architecture rather than specific to SM and AM. We keep the network skeleton fixed and replace SM and AM throughout S1--S3 and the bottleneck of the unpruned network with global attention, Gated DeltaNet (GDNet)~\cite{yang2025gated}, $3\times3\times3$ depthwise convolution (DWConv), or plain $3\times3\times3$ convolution (Conv). Across all five networks, the bottleneck has the highest ARD and lowest CKA, whereas S2/S3 have the smallest ARD and high CKA (Fig.~\ref{fig.7}(a,b)).

ARD consistently distinguishes S1 from S2/S3: S1's score is 1.7--20 times the larger of the S2 and S3 scores, with the largest ratio in the unpruned network. The bottleneck's ARD exceeds that of every encoder stage by at least fivefold. By contrast, encoder CKA values cluster near one and do not consistently identify S1 as less redundant than both S2 and S3; for DWConv, S1 has slightly higher CKA than S2.

This stage ordering holds in all nine runs of the unpruned network across three datasets and three training seeds (Fig.~\ref{fig.7}(c)). On every dataset, bottleneck ARD exceeds that of every encoder stage by more than an order of magnitude. S1 exceeds S2/S3 by a clear margin on AMOS and ACDC and by a smaller margin on AbdomenCT-1K.

\begin{table*}[!tb]
\centering
\caption{
Token-mixer replacement in SlimFormer++.
Ours gives absolute values (Dice and Jaccard in \%, HD95 and ASD in mm); other rows give changes relative to Ours.
$\Delta$P: parameter change (M); $\Delta$G: GFLOP change (\%, averaged over datasets).
}
\label{tab:mixer_replacement}
\footnotesize
\renewcommand{\arraystretch}{1.12}
\newcommand{\mixertablebody}{%
\toprule
& & &
\multicolumn{4}{c}{AbdomenCT-1K} &
\multicolumn{4}{c}{AMOS} &
\multicolumn{4}{c}{ACDC} \\
\cmidrule(lr){4-7}
\cmidrule(lr){8-11}
\cmidrule(lr){12-15}
Mixer &
$\Delta$P$\downarrow$ &
$\Delta$G$\downarrow$ &
Dice$\uparrow$ & Jaccard$\uparrow$ & HD95$\downarrow$ & ASD$\downarrow$ &
Dice$\uparrow$ & Jaccard$\uparrow$ & HD95$\downarrow$ & ASD$\downarrow$ &
Dice$\uparrow$ & Jaccard$\uparrow$ & HD95$\downarrow$ & ASD$\downarrow$ \\
\midrule
\rowcolor{oursrow}
Ours &
-- & -- &
95.62 & 91.83 & 2.84 & 0.67 & 89.76 & 84.14 & 11.36 & 6.38 & 91.56 & 85.05 & 3.70 & 0.69 \\
\midrule
\multicolumn{15}{l}{\textit{Encoder: alternative mixers activated throughout S1--S3}} \\
Attention &
+0.34 & +37.9 &
$-$0.10 & $-$0.17 & $-$0.01 & +0.04 & $-$0.28 & $-$0.40 & $-$0.10 & $-$0.29 & $-$0.20 & $-$0.34 & +0.84 & +0.09 \\
GDNet &
+0.75 & +21.9 &
$-$0.30 & $-$0.49 & +0.69 & +0.10 & $-$0.27 & $-$0.40 & $-$0.70 & $-$0.89 & $-$0.16 & +0.08 & +0.24 & +0.01 \\
DWConv &
+0.03 & +6.8 &
$-$0.11 & $-$0.18 & +0.52 & $-$0.15 & $-$0.25 & $-$0.36 & $-$0.88 & $-$0.87 & +0.05 & +0.04 & +0.30 & 0.00 \\
Conv &
+2.57 & +23.8 &
$-$0.10 & $-$0.18 & +0.38 & +0.06 & $-$0.51 & $-$0.70 & $-$0.49 & $-$0.40 & $-$0.05 & $-$0.11 & +0.84 & +0.10 \\
\midrule
\multicolumn{15}{l}{\textit{Bottleneck: AM replaced}} \\
Attention &
+0.57 & +0.4 &
$-$0.16 & $-$0.26 & +1.00 & +0.14 & $-$0.53 & $-$0.72 & $-$0.10 & $-$0.37 & $-$0.03 & $-$0.05 & +0.68 & +0.05 \\
GDNet &
+0.88 & +0.6 &
$-$0.09 & $-$0.17 & +0.29 & +0.08 & $-$0.60 & $-$0.80 & +1.50 & +0.98 & $-$0.17 & $-$0.25 & +0.78 & +0.07 \\
DWConv &
$-$0.006 & $-$0.1 &
$-$0.07 & $-$0.12 & $-$0.07 & 0.00 & $-$0.41 & $-$0.56 & +1.20 & +0.31 & $-$0.42 & $-$0.62 & $-$0.10 & +0.12 \\
Conv &
+3.96 & +1.3 &
$-$0.12 & $-$0.17 & +0.48 & +0.07 & $-$0.50 & $-$0.66 & +0.89 & +0.65 & $-$0.20 & $-$0.31 & +0.08 & +0.04 \\
\bottomrule%
}
\newsavebox{\mixertablebox}
\setlength{\tabcolsep}{0pt}
\sbox{\mixertablebox}{\begin{tabular}{lrr*{3}{rrrr}}\mixertablebody\end{tabular}}
\setlength{\tabcolsep}{\dimexpr(\textwidth-\wd\mixertablebox)/30\relax}
\begin{tabular}{lrr*{3}{rrrr}}\mixertablebody\end{tabular}
\end{table*}

\subsection{Token-Mixer Replacement}
\label{sec 4.4}

We evaluate the four alternative mixers from Section~\ref{sec 4.3.2} within SlimFormer++ by activating them throughout S1--S3 (changing both mixer type and placement) or replacing AM at the bottleneck (changing only mixer type). Table~\ref{tab:mixer_replacement} reports changes relative to SlimFormer++.

\textbf{Encoder.}
Activating mixers throughout S1--S3 increases cost without consistent accuracy gains. Attention adds the most computation (+37.9\% GFLOPs), whereas Conv adds the most parameters (+2.57M). None consistently improves Dice, and HD95 improvements are largely confined to AMOS. These results favor the selective SM encoder, consistent with the small S2/S3 contributions in Fig.~\ref{fig.7}(b,c).

\textbf{Bottleneck.}
None of the evaluated replacements consistently improves Dice across the datasets, suggesting that the proposed bottleneck design has already achieved a balance between representational capacity and computational efficiency. DWConv saves only 0.006M parameters and 0.1\% GFLOPs but lowers AMOS Dice by 0.41 pp and increases HD95 by 1.20 mm. Since AM reduces to DWConv when all $\beta_c=0$ (Section~\ref{sec 3.3.2}), this comparison supports the benefit of its remote-context augmentation at little additional cost. The remaining alternatives add 0.57--3.96M parameters, with GFLOP increases of at most 1.3\% at the low-resolution bottleneck.

\begin{table*}[!t]
\centering
\caption{Quantitative comparison with state-of-the-art 3D medical image segmentation methods on eight public datasets. Dice is in \% ({\unboldmath$\uparrow$}) and HD95 in mm ({\unboldmath$\downarrow$}). \textbf{Bold}: best; \underline{underlined}: second-best distinct value for each dataset and metric.}
\label{tab:main_results}
\scriptsize
\setlength{\tabcolsep}{1.6pt}
\renewcommand{\arraystretch}{1.1}
\begin{tabular}{@{}l r@{\hspace{9pt}}r@{\hspace{2.2pt}}l *{7}{@{\hspace{3pt}}r@{\hspace{2.2pt}}l}@{}}
\toprule
\multirow{2}{*}{Model} & Params
& \multicolumn{2}{c}{AbdomenCT-1K}
& \multicolumn{2}{c}{ACDC}
& \multicolumn{2}{c}{AMOS}
& \multicolumn{2}{c}{CHAOS}
& \multicolumn{2}{c}{RESECT}
& \multicolumn{2}{c}{MMWHS}
& \multicolumn{2}{c}{VESSEL12}
& \multicolumn{2}{c}{Leg3DUS} \\
\cmidrule(l{3pt}r{1pt}){3-4} \cmidrule(l{3pt}r{1pt}){5-6} \cmidrule(l{3pt}r{1pt}){7-8}
\cmidrule(l{3pt}r{1pt}){9-10} \cmidrule(l{3pt}r{1pt}){11-12} \cmidrule(l{3pt}r{1pt}){13-14}
\cmidrule(l{3pt}r{1pt}){15-16} \cmidrule(l{3pt}r{1pt}){17-18}
& (M) & Dice$\uparrow$ & HD95$\downarrow$ & Dice$\uparrow$ & HD95$\downarrow$ & Dice$\uparrow$ & HD95$\downarrow$ & Dice$\uparrow$ & HD95$\downarrow$ & Dice$\uparrow$ & HD95$\downarrow$ & Dice$\uparrow$ & HD95$\downarrow$ & Dice$\uparrow$ & HD95$\downarrow$ & Dice$\uparrow$ & HD95$\downarrow$ \\
\midrule
\multicolumn{18}{@{}l}{\textit{Representative models}} \\
nnUNet~\cite{isensee2021nnu}\venue{Nat.\,Methods'21} & 30.8 & 94.73 & 4.59 & 90.06 & 4.71 & 83.61 & 21.01 & 79.66 & 17.61 & \underline{68.12} & 6.98 & 87.63 & 6.56 & 99.51 & \underline{0.74} & 81.37 & 13.10 \\
TransUNet~\cite{chen2024transunet}\venue{MedIA'24} & 151.1 & \underline{95.42} & \underline{3.34} & 90.63 & 4.74 & 89.64 & 11.81 & 80.92 & 17.54 & 66.56 & 8.13 & 88.93 & 6.68 & 99.55 & \underline{0.74} & 83.37 & \textbf{11.41} \\
SlimFormer backbone~\cite{hong2025slimformer} & 69.8 & 94.39 & 76.52 & 88.12 & 4.46 & 84.37 & 17.19 & 79.38 & 18.06 & 64.74 & 5.98 & 88.04 & 7.51 & 98.33 & 10.40 & 76.44 & 24.51 \\
nnWNet3D~\cite{zhou2025nnwnet}\venue{CVPR'25} & 53.2 & 95.11 & 3.95 & 91.47 & 3.97 & 87.50 & 13.21 & \underline{83.70} & \underline{12.78} & 67.80 & 6.64 & 90.00 & 6.36 & \textbf{99.60} & \textbf{0.71} & 82.50 & 19.83 \\
SegMambaV2~\cite{xing2025segmamba}\venue{TMI'26} & 138.8 & 92.30 & 10.00 & \underline{91.50} & \underline{3.93} & 83.50 & 17.46 & 80.90 & 14.40 & 67.93 & 7.31 & 87.80 & 6.70 & 99.56 & \underline{0.74} & 81.60 & 23.18 \\
\midrule
\multicolumn{18}{@{}l}{\textit{Pruned and lightweight models}} \\
nnUNet-Pruned~\cite{isensee2021nnu} & 18.8 & 94.19 & 5.62 & 89.39 & 5.89 & 83.87 & 18.41 & 80.70 & 17.61 & 66.36 & 7.16 & 85.21 & 9.99 & 99.46 & \textbf{0.71} & 82.61 & 11.59 \\
TransUNet-Pruned~\cite{chen2024transunet} & 62.2 & 95.19 & 4.85 & 90.32 & 4.51 & 88.81 & 11.65 & 81.35 & 17.52 & 67.20 & 7.67 & 88.55 & 7.50 & 99.54 & \underline{0.74} & 83.75 & 12.22 \\
SlimFormer~\cite{hong2025slimformer}\venue{MICCAI'25} & 17.4 & 94.04 & 9.92 & 87.37 & 6.22 & 84.08 & 18.99 & 79.67 & 18.22 & 65.83 & \underline{5.81} & 87.76 & 6.06 & 97.88 & 19.28 & 75.62 & 26.36 \\
CROWn~\cite{huang2026crown}\venue{CVPR'26} & 23.9 & 94.80 & 4.44 & 90.40 & 4.19 & 87.50 & 12.93 & 81.40 & 16.49 & 67.70 & 7.79 & \textbf{90.90} & \textbf{5.14} & \underline{99.58} & \textbf{0.71} & 81.70 & 17.21 \\
PnPNet~\cite{you2026towards}\venue{MedIA'26} & 27.4 & 70.00 & 37.77 & 77.30 & 35.18 & 81.20 & 21.67 & 82.00 & 13.21 & 66.61 & 7.85 & 82.00 & 10.93 & 96.60 & 8.60 & 80.30 & 19.67 \\
MedNeXt~\cite{roy2023mednext}\venue{MICCAI'23} & 11.0 & 95.20 & 4.13 & 90.60 & 4.66 & 88.40 & 11.69 & 82.60 & 16.40 & 67.79 & 7.56 & 87.10 & 6.46 & \textbf{99.60} & \textbf{0.71} & \textbf{84.00} & 12.08 \\
TfeNet~\cite{WU2026103882}\venue{MedIA'26} & \underline{8.7} & 94.20 & 79.69 & 88.38 & 5.11 & \textbf{90.01} & \textbf{11.13} & 80.75 & 15.84 & 64.25 & 26.18 & 76.54 & 33.06 & 99.37 & 0.88 & 78.10 & 32.64 \\
\midrule
\rowcolor{oursrow}
\textbf{SlimFormer++ (Ours)} & \textbf{5.98} & \textbf{95.62} & \textbf{2.84} & \textbf{91.56} & \textbf{3.70} & \underline{89.76} & \underline{11.36} & \textbf{83.88} & \textbf{10.06} & \textbf{68.29} & \textbf{5.71} & \underline{90.18} & \underline{6.03} & \textbf{99.60} & \textbf{0.71} & \underline{83.80} & \underline{11.58} \\
\bottomrule
\end{tabular}
\end{table*}

\subsection{Comparison with State-of-the-Art Methods}
\label{sec 4.5}

Table~\ref{tab:main_results} compares SlimFormer++ with five representative models, three block-pruned variants, and four lightweight models on eight datasets. With the fewest parameters (5.98M), SlimFormer++ ranks first or second in all 16 dataset--metric pairs and achieves the highest mean Dice (87.84\%) and lowest mean HD95 (6.50~mm), improving on the respective next-best models by 0.63~pp and 1.46~mm.

Against the representative models, SlimFormer++ matches or exceeds their Dice on every dataset while using 5.2--25.3 times fewer parameters. Block pruning removes 39.0--75.1\% of the parameters of nnUNet, TransUNet, and the SlimFormer backbone with mean Dice decreases of 0.04--0.37~pp, but the resulting variants still require 17.4--62.2M parameters. SlimFormer++ achieves higher Dice than all three pruned variants on every dataset, with mean gains of 1.00--3.81~pp, and higher mean Dice than all four lightweight models.

\begin{figure}[!tb]
    \centering
    \includegraphics[width=\columnwidth]{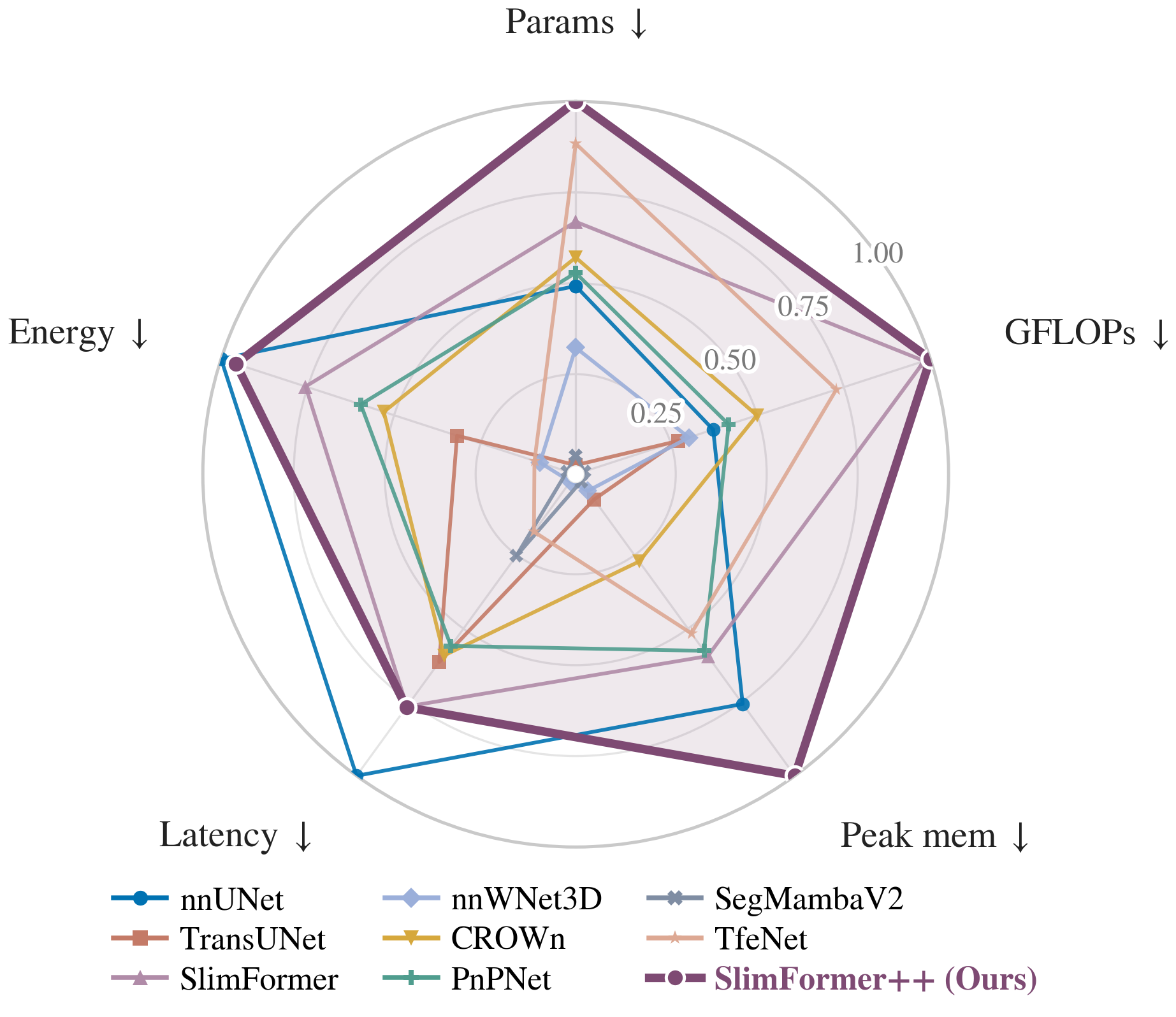}
    \caption{
    Efficiency profiles of nine models: parameters (M), GFLOPs, latency (ms), peak memory (GB), and energy (J). Values are min--max normalized in logarithmic space and \textbf{inverted}, so larger radii indicate lower costs within the displayed comparison set.
    }
    \label{fig.8}
\end{figure}

\begin{figure*}[!t]
    \centering
    \includegraphics[width=\textwidth]{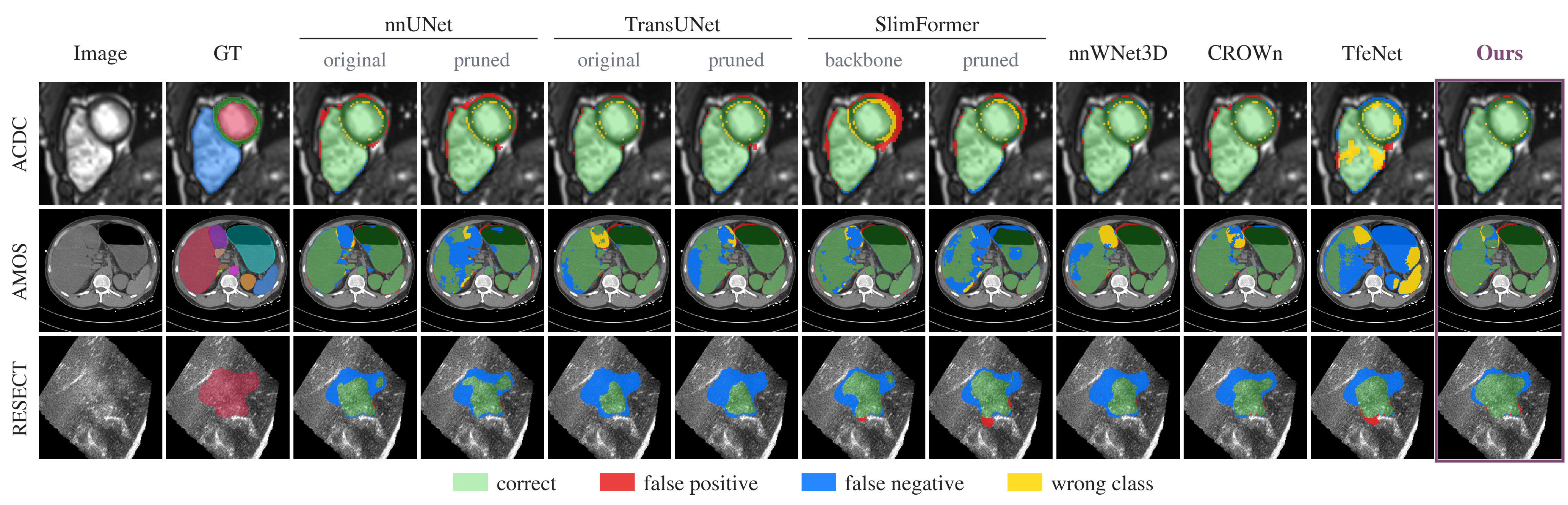}
    \caption{Qualitative comparison on three representative datasets covering CT, MRI, and ultrasound (rows, ordered as in Table~\ref{tab:dataset_summary}).}
    \label{fig.9}
\end{figure*}

\subsection{Computational Efficiency}
\label{sec 4.6}

Fig.~\ref{fig.8} compares the parameter count, GFLOPs, peak inference memory, latency, and energy of SlimFormer++ with those of eight comparison models, all measured for one forward pass of a $96\times96\times96$ patch on a single GH200 GPU.

Among the models shown in the radar chart, SlimFormer++ has the fewest parameters, the lowest computational cost, and the lowest peak memory usage. It requires 189.3~GFLOPs, 4.7\% fewer than the next-lowest model on this axis, SlimFormer (198.6~GFLOPs), while using 65.6\% fewer parameters and achieving a mean Dice 3.81~pp higher (87.84\% versus 84.03\%). Its peak memory of 0.43~GB is 29.5\% lower than that of nnUNet (0.61~GB), the next-lowest model.

For latency and energy consumption, nnUNet ranks first (7.33~ms and 1.93~J), followed by SlimFormer++ (12.99~ms and 2.08~J per volume), although nnUNet requires 1,065.8~GFLOPs, 5.6 times that of SlimFormer++. The latency gap reflects memory-access overhead in the SM scan and AM remote gathers, whose kernels are less optimized than dense cuDNN convolutions. Fused kernels for scan and gather operations may help translate lower GFLOPs into faster inference.

\subsection{Qualitative Comparison}
\label{sec 4.7}

Fig.~\ref{fig.9} compares predictions on one selected slice per dataset, with overlays distinguishing correct labels, false positives, false negatives, and class confusions. In RESECT, all models under-segment the tumor; SlimFormer++ recovers most of it and misses mainly thin bands along its upper boundary, whereas both TransUNet variants and nnWNet3D miss most of the tumor, and the pruned SlimFormer and TfeNet add false positives below it. In AMOS, SlimFormer++ leaves only small missed regions.

\section{Conclusion}

Redundancy analysis guided block removal and stage-wise token-mixer allocation for 3D medical image segmentation. CKA identified redundant blocks and informed the hybrid macro architecture, whereas ARD located the strongest mixer contribution at the bottleneck and separated S1 from S2/S3 across the tested mixers, motivating AM at the bottleneck, SM in S1, and residual channel MLPs in S2/S3. Across eight public CT, MRI, and ultrasound datasets, SlimFormer++ achieved the highest mean Dice (87.84\%) with 5.98 million parameters. Its latency still exceeds that of nnUNet despite lower GFLOPs (Section~\ref{sec 4.6}); future work will incorporate measured latency into the design criterion and develop hardware-efficient mixer kernels.

\bibliographystyle{IEEEtran}
\newcommand{\BIBdecl}{\setlength{\baselineskip}{0.95\baselineskip}}
\bibliography{refs}

\end{document}